\UseRawInputEncoding
\pdfoutput=1

\documentclass[11pt]{article}
\usepackage{caption}
\usepackage{subcaption}
\usepackage{svg}
\usepackage[final]{acl}
\usepackage{times}
\usepackage{latexsym}
\usepackage{amsmath}
\usepackage{amsfonts}
\usepackage[T1]{fontenc}

\usepackage[utf8]{inputenc}
\usepackage{booktabs}
\usepackage{microtype}
\usepackage{graphicx}
\usepackage{dashrule}
\usepackage{amsfonts,amssymb}
\usepackage{arydshln} 
\usepackage{booktabs}
\usepackage{multirow}
\usepackage{microtype}

\title{Tracing Origins: Coreference-aware Machine Reading Comprehension}

\author{
Baorong Huang\textsuperscript{1,\#}, 
Zhuosheng Zhang\textsuperscript{2,3,\#},
Hai Zhao\textsuperscript{2,3,\thanks{\; Corresponding author. \# Equal contribution. This work was supported in part by the Key Projects of National Natural Science Foundation of China
under Grants U1836222 and 61733011.}} \\
 $^1$Institute of Corpus Studies and Applications, Shanghai International Studies University \\
  $^2$Department of Computer Science and Engineering,  Shanghai Jiao Tong University
   \\ $^3$Key Laboratory of Shanghai Education Commission for Intelligent Interaction \\
and Cognitive Engineering, Shanghai Jiao Tong University \\
  \texttt{0214101730@shisu.edu.cn,zhangzs@sjtu.edu.cn,zhaohai@cs.sjtu.edu.cn} 
  }
\date{}

\begin{document}

\maketitle
\begin{abstract}
Machine reading comprehension is a heavily-studied research and test field for evaluating new pre-trained language models (PrLMs) and fine-tuning strategies, and recent studies have enriched the pre-trained language models with syntactic, semantic and other linguistic information to improve the performance of the models. In this paper, we imitate the human reading process in connecting the anaphoric expressions and explicitly leverage the coreference information of the entities to enhance the word embeddings from the pre-trained language model, in order to highlight the coreference mentions of the entities that must be identified for coreference-intensive question answering in QUOREF, a relatively new dataset that is specifically designed to evaluate the coreference-related performance of a model. We use two strategies to fine-tune a pre-trained language model, namely, placing an additional encoder layer after a pre-trained language model to focus on the coreference mentions or constructing a relational graph convolutional network to model the coreference relations. We demonstrate that the explicit incorporation of coreference information in the fine-tuning stage performs better than the incorporation of the coreference information in pre-training a language model. 
\end{abstract}

\section{Introduction}
Machine reading comprehension (MRC), a task that automatically identifies one or multiple words from a given passage as the context to answer a specific question for that passage, is widely used in information retrieving, search engines, and dialog systems. Several datasets on MRC that limit the answer to one single word or multiple words from the passage are introduced, including TREC \cite{Voorhees2003}, SQuAD \cite{rajpurkar-etal-2018-know}, NewsQA \cite{trischler2016newsqa}, SearchQA \cite{dunn2017searchqa}, and QuAC \cite{Choi2018}, and intensive efforts were made to build new models that surpass the human performance on these datasets, including the pre-trained language models \cite{devlin-etal-2019-bert, Liu2019RoBERTaAR, yang2020xlnet} or the ensemble models that outperform the human, in particular on SQuAD \cite{lan2020albert, yamada-etal-2020-luke, zhang2020retrospective}. More challenging datasets are also introduced, which require several reasoning steps to answer \cite{yang-etal-2018-hotpotqa,qi-etal-2021-answering},  the understanding of a much larger context \cite{kocisky-etal-2018-narrativeqa} or the understanding of the adversarial content and numeric reasoning \cite{Dua2019DROP}.   

    \begin{table}
        \centering\small
        \begin{tabular}{|p{7.2cm}|}
        \hline
       \textcolor{blue}{Context: }  \textit{\textbf{Frankie Bono}, a mentally disturbed hitman from 
        Cleveland, comes back to his hometown in New York City during Christmas week to kill a middle-management mobster, Troiano. ...\textcolor{teal}{First \textbf{\underline{he}} follows his target to select the best possible location, but opts to wait until Troiano isn't being accompanied by his bodyguards.} ... Losing his nerve, \textbf{Frankie} calls up his employers to tell them he wants to quit the job. \textcolor{teal}{Unsympathetic, the supervisor tells him \textbf{\underline{he}} has until New Year's Eve to perform the hit. } }
        \\
       \hline
       \textcolor{blue}{Question:} \textit{What is the first name of the person who has until New Year's Eve to perform a hit?} \textcolor{blue}{Answer:} \textcolor{teal}{\underline{he}} -\textgreater \textcolor{teal}{\underline{Frankie}} \\
        \hline
       \textcolor{blue}{Question:}   \textit{What is the first name of the person who follows their target to select the best possible location?} \textcolor{blue}{Answer:} \textcolor{teal}{\underline{he}} -\textgreater \textcolor{teal}{\underline{Frankie}} \\
        \hline
        
        \end{tabular}
        \caption{An example from QUOREF: coreference resolution is required to extract the correct answer. We highlight the supporting text \textcolor{teal}{in teal color} and the related deictic expressions in \textbf{bold}.}
        \label{tab:my_label}
    \end{table}

Human texts, especially long texts, are abound in deictic and anaphoric expressions that refer to the entities in the same text. These deictic and anaphoric expressions, in particular, constrain the generalization of the models trained without explicit awareness of the coreference. The QUOREF dataset \cite{Dasigi2019Quoref} is specifically designed to validate the performance of the models in coreferential reasoning, in that ``78\% of the manually analyzed questions cannot be answered without coreference'' \cite{Dasigi2019Quoref}. The example in Table 1 shows that the answers to the two questions cannot be directly retrieved from the sentences because the word in the corresponding sentence of the context is 
an anaphoric pronoun \textit{\textbf{he}}, and to obtain
the correct answers, tracing of its antecedent \textit{\textbf{Frankie}} is required. The reasoning in coreference resolution is required to successfully complete the task in machine reading comprehension in the SQuAD-style QUOREF dataset.

Pre-trained language models, including BERT \cite{devlin-etal-2019-bert}, RoBERTa \cite{Liu2019RoBERTaAR} and XLNet \cite{Yang2019XLNetGA}, that are trained through self-supervised language modeling objectives like masked language modeling, perform rather poorly in the QUOREF dataset. 
We argue that the reason for the poor performance is that those pre-trained language models do learn the background knowledge for coreference resolution but may not learn adequately the coreference information required for the coreference-intensive reading comprehension tasks. In the human reading process, as shown in the empirical study of  first-year English as a second language students during the reading of expository texts, ``anaphoric resolution requires a reader to perform a text-connecting task across textual units by successfully linking an appropriate antecedent (among several prior antecedents) with a specific anaphoric referent'' and ``students who were not performing well academically were not skilled at resolving anaphors'' \cite{Pretorius2005} and the direct instruction on anaphoric resolution elevated the readers' comprehension of the text \cite{Baumann1986}. In addition, the studies on anaphor resolution in both adults using eye movement studies \cite{Duffy1990,Gompel04antecedenttypicality} and children \cite{HollyJoseph2015} evidenced a two-stage model of anaphor resolution proposed by Garrod and Terras\cite{Garrod2000}. The first stage is “an initial lexically driven, context-free stage known as bonding, whereby a link between the anaphor and a potential antecedent is made, followed by a later process known as resolution, which resolves the link with respect to the overall discourse context” \cite{HollyJoseph2015}. The pre-trained language models only capture the semantic representations of the words and sentences, without explicitly performing such text-connecting actions in the specific coreference-intensive reading comprehension task, thus they do not learn adequate knowledge to solve the complex coreference reasoning problems.

Explicitly injecting external knowledge such as linguistics and knowledge graph entities, has been shown effective to broaden the scope of the pre-trained language models' capacity and performance, and they are often known as X-aware pre-trained language models \cite{Zhang2020,Liu2020,Kumar2021}. It is plausible that we may imitate the anaphoric resolution process in human's two-stage reading comprehension of coreference intensive materials and explicitly make the text-connecting task in our fine-tuning stage as the second stage in the machine reading comprehension.

As an important tool that captures the anaphoric relationship between words or phrases, coreference resolution that clusters the mentions of the same entity within a given text is an active field in natural language processing \cite{chen-etal-2011-unified, Sangeetha2012, Huang2019,joshi-etal-2020-spanbert, kirstain-etal-2021-coreference}, with neural networks taking the lead in the coreference resolution challenges. The incorporation of the coreference resolution results in the pre-training to obtain the coreference-informed pre-trained language models, such as CorefBERT and  CorefRoBERTa \cite{ye-etal-2020-coreferential}, has shown positive improvements on the QUOREF dataset, a dataset that is specially designed for measuring the models' coreference capability, but the performance is still considerably below the human performance.

In this paper, we make a different attempt to leverage the coreference resolution knowledge and complete the anaphoric resolution process in reading comprehension. We propose a fine-tuned coref-aware model that directly instructs the model to learn the coreference information\footnote{Our codes are publicly available at
\url{https://github.com/bright2013/CorefAwareMRC}.}. Our model can be roughly divided into three major components: 1) pre-trained language model component. We use the contextualized representations from the pre-trained language models as the token embeddings for the downstream reading comprehension tasks. 2) coreference resolution component. NeuralCoref, an extension to the spaCy, is applied here to extract the mention clusters from the context. 3) coreference enrichment component. We apply three methods in incorporating the coreference knowledge: additive attention enhancement, multiplication attention enhancement,  and relation-enhanced graph-attention network + fusing layer. 

In this paper, we show that by simulating the human behavior in explicitly connecting the  anaphoric expressions to the antecedent entities and infusing the coreference knowledge our model can surpass that of the pre-trained coreference language models on the QUOREF dataset.


\begin{figure*}[htp]
  \centering
  \includegraphics[width=\textwidth]{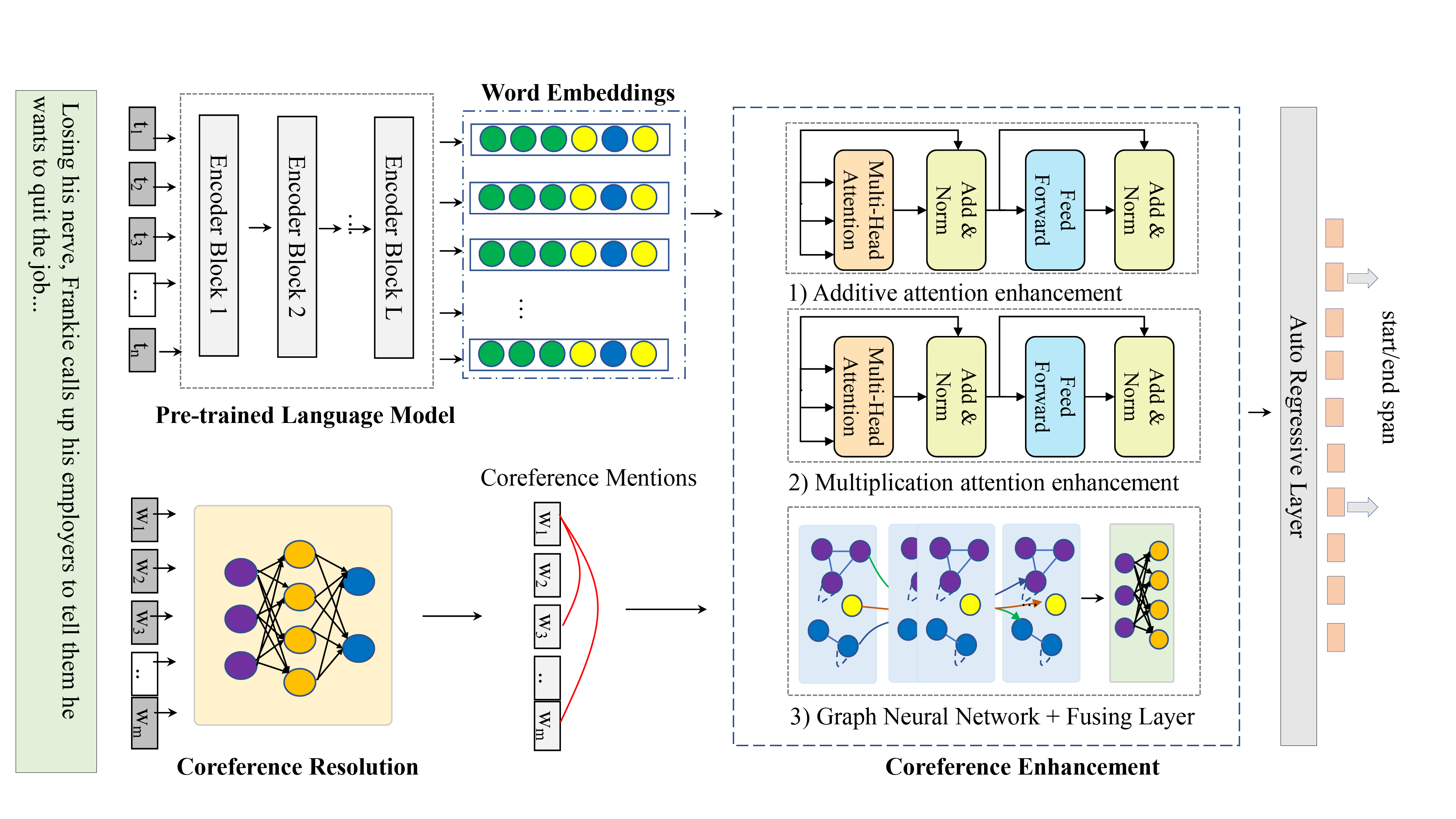}
  \caption{Coref-aware fine-tuning for machine reading comprehension. The text is tokenized and fed into a pre-trained language model to obtain the embeddings, and into a coreference resolution model to obtain coreference information. Both the embeddings and the coreference information are used in the fine-tuning stage to 1) enhance cross attentions with additive operations; 2) enhance cross attentions with multiplication operations, or; 3) construct a coreference graph neural network with the coreference relations as edges. }
  \label{fig:model}
\end{figure*}

\section{Background and Related Work}
\subsection{Models and Training Strategies}
Recent studies on machine reading comprehension mainly rely on the neural network approaches. Before the prevalence of the pre-trained language models, the main focus was to guide and fuse the attentions between questions and paragraphs in their models, in order to gain better global and attended representation \cite{huang2018fusionnet,Minghao2018-570,wang-etal-2018-multi-granularity}. 

After the advent of the BERT \cite{devlin-etal-2019-bert}, there were two trends in solving the machine reading comprehension. The first trend was to develop better pre-trained language models that captured the representation of contexts and questions \cite{Liu2019RoBERTaAR,yang2020xlnet,lewis-etal-2020-bart}, and more datasets on question answering were introduced to increase the difficulty in this task,  including NewsQA \cite{trischler2016newsqa}, SearchQA \cite{dunn2017searchqa}, QuAC \cite{Choi2018}, HotpotQA \cite{yang-etal-2018-hotpotqa}, NarrativeQA \cite{kocisky-etal-2018-narrativeqa}, DROP \cite{Dua2019DROP}, and BeerQA \cite{qi-etal-2021-answering}. 


However, the raw pre-trained language models, being deprived of the in-domain knowledge, the structures and the reasoning capabilities required for the datasets, often perform unsatisfactorily in the hard datasets, being significantly below the human performance. Efforts had been made to boost the model performance by enriching the pre-trained language models with specific syntactic information \cite{ye-etal-2020-coreferential} or semantic information. Another trend was to fine-tune the pre-trained language model and added additional layers to incorporate task-specific information for better representation, in particular, the coreference information \cite{ouyang-etal-2021-dialogue,liu2021coreferenceaware}. For some questions that have multi-span answers, in other words, a single answer contains two or more discontinuous entities in the context, the BIO (B denotes the start token of the span; I denotes the subsequent tokens and O denotes tokens outside of the span) tagging mechanism is used to identify these answers and improve the model performance \cite{segal-etal-2020-simple}.

Recent studies also explored the possibilities of prompt-based learning in machine reading comprehension, including a new pre-training scheme that changed the question answering into a few-shot span selection model \cite{ram-etal-2021-shot} and a new model that fine-tuned the prompts with knowledge \cite{Chen2021KnowPromptKP}. The performance of the models using prompt-based learning is significantly higher than the baseline models, but is still below that of the fine-tuned models \cite{Chen2021KnowPromptKP}.

\subsection{Graph Neural Network in Machine Reading Comprehension}
Graph neural network (GNN) captures the relations among the entities in the text by modeling the entities as nodes in the graph and learning the weights via the message passing between the nodes of the graph \cite{Kipf2017SemiSupervisedCW, veli2018graph}. As the dependencies in the natural language text,  the relations among entities and knowledge-base triples can be relatively easily modeled in a graph structure, graph neural networks 
are used for numeric reasoning \cite{ran-etal-2019-numnet}, for multi-document question answering by connecting mentions of candidate answers \cite{de-cao-etal-2019-question}, and for multi-hop reasoning by adding the edges with co-occurrence relations\cite{ qiu-etal-2019-dynamically}, or with contextual sentences as embeddings \cite{Tuarticle2020}, or with a hierarchical paragraph-sentence-entity graph \cite{ fang-etal-2020-hierarchical}, but none of them had attempted to connect the anaphoric expressions and their antecedents as a coreference resolution strategy in a graph neural network for machine reading comprehension.

\section{Coreference-aware Machine Reading Comprehension}

Our model, inspired by the anaphoric connecting behavior in the human reading comprehension process,  consists of four parts, namely, a pre-trained language model, a coreference resolution component, a graph encoder and a fusing layer. Context in the machine reading comprehension task is first processed by a coreference resolution model to identify the underlying coreference clusters, which are formed by dividing the entities and anaphoric expressions in the context into disjoint groups on the principle that the mentions of the same entity should be in the same group. Then we use the coreference clusters to construct a coreference matrix that labels each individual cluster and identifies each element in the same cluster with the same cluster number. Meanwhile, the context is tokenized by the tokenizer defined in the pre-trained language model and the embeddings for each token are retrieved from that model. We propose three methods for connecting the anaphoric expressions and their antecedent entity: 1) adding the coreference matrix with each attention head in the additional coreference encoder layer; 2) multiplying the coreference matrix with each attention head in the additional coreference encoder layer; 3) 
constructing a graph neural network based on the coreference matrix with the edges corresponding to the coreference relations and then fusing the graph representation in the graph neural network with the embeddings of the context, as shown in Figure \ref{fig:model}. The final representations from either one of the three methods are fed into the classifier to calculate the start/end span of the question.

 \begin{figure*}[ht]
  \centering
  \includegraphics[width=16.25cm, height=4.00cm]{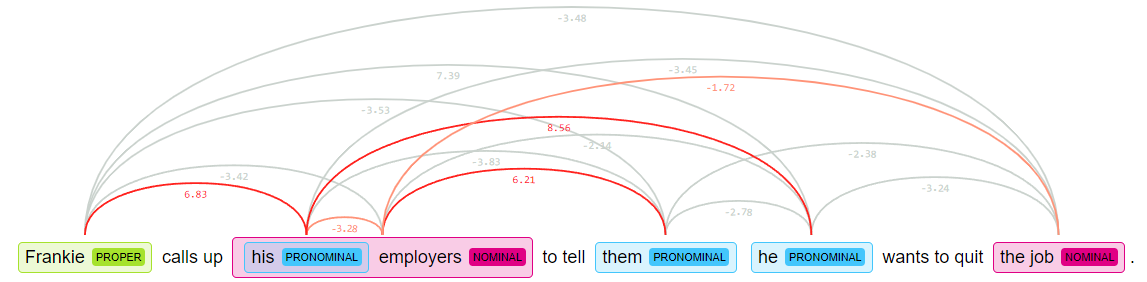}
  \caption{Coreference resolution: the red curves connecting the mentions of the same entity and marking the coreference relations. \protect\footnotemark}
  \label{fig:coref}
\end{figure*}

\footnotetext{Image generated from https://huggingface.co/coref/}

\subsection{Coreference Resolution}
 Coreference resolution is the process that identifies all the expressions of the same entity in the text, clusters them together as coreference clusters, and locates their spans. For example, after coreference resolution for the text \textit{Losing his nerve, Frankie calls up his employers to tell them he wants to quit the job.}, we obtained two mention clusters \textit{[Frankie: [his, Frankie, his, he], his employers: [his employers, them]]}, where \textit{Frankie} is the head entity and \textit{his, Frankie, his, he} are all the expressions referring to this entity,  as shown in Figure \ref{fig:coref}.

As pre-trained language models use subwords in their tokenization and the coreference resolution uses word in the tokenization, a mapping is required to establish the relations. For the input sequence \textit{ $X = \{x_1,...x_n\}$} of length n, the words \textit{ $W = \{w_1,...,w_m\}$ } obtained from the coreference tokenization are mapped to the corresponding subwords (tokens) \textit{ $T = \{t_1,..., t_k\}$ } from the tokenizer in the pre-trained language model, with one word contains one or more than one subwords.
Then we constructed a coreference array with the following rule:
\begin{equation}
coref(i) = \left \{
\begin{aligned} 
0& \text{ $ \rm \:if\: tokens[i] \notin S_m$}, \\
n& \text{ $ \rm \:if\: tokens[i] \in S_m$},
\end{aligned}
 \right .
\end{equation}
where \textit{i} is the position of the token in the token array, $ S_m $ is an array of all words in the coreference mention clusters, \textit{n} is the sequence number of the mention cluster and $n \geq 1$. Tokens in the same mention cluster have the same sequence number \textit{n} in the coreference array.

\subsection{Graph Neural Network}
\begin{figure}[htbp]
  \centering
  \includegraphics[width=0.4\textwidth]{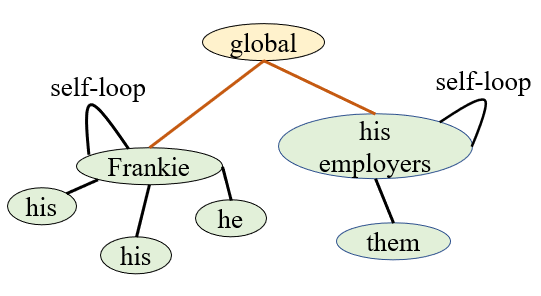}
  \caption{Coreference graph. We connect the entities with their coreference mentions to form a graph, and the nodes are connected to the global node to form global representations.}
  \label{fig:corefgraph}
\end{figure}

We use the standard relational graph convolutional network (RGCN) \cite{sejrschlichtkrull2018modeling} to obtain the graph representation of the context enriched with coreference information. We use the coreference matrix and the word embeddings to construct a directed and labeled graph $\mathcal{G = (V,E,R)}$, with nodes (subwords) ${v_i \in \mathcal{V}}$, edges(relations) ${(v_i, r, v_j)) \in \mathcal{E}} $, where ${r \in \mathcal{R}} $ is one of the two relation types (1 indicates coreference relation and self-loop; 2 indicates global relation), as shown in Figure \ref{fig:corefgraph} .

The constructed graph is then fed into the RGCN, with the differentiable message passing and the basis decomposition to reduce model parameter size and prevent overfitting:
\begin{equation}
\begin{aligned}
& h_i^{l+1} = \sigma \big(W^{(l)}_0h^{(l)}_i + \sum_{r \in R}\sum_{j \in N^r_i}\frac{1}{c_{i,r}}W_r^{(l)}h^{(l)} \big),  \\
& W^{(l)}_r = \sum_{b=1}^Ba_{rb}^{(l)}V_b^{(l)} ,
\end{aligned}
\end{equation}
where $N^r_i$ denotes the set of neighbor indices of node i under the relation ${r \in \mathcal{R}}$, $c_{i,r}$ is the normalization constant, and $W^{(l)}_r$ is a linear combination of basis transformation $V_b^{(l)}$ with coefficient $a_{rb}^{(l)}$.
\subsection{Coreference-enhanced Attention}
In addition to the Graph Neural Network method, we also explore the possibility of using the self-attention mechanism \cite{Vaswani2017} to explicitly add an encoder layer and incorporate the coreference information into the attention heads of that layer, so as to guide the model to identify the mentions in the cluster as the same entity.

We use two methods to fuse the coreference information and the original embeddings from the pre-trained language model: additive attention fusing and dot product attention fusing (multiplication). Given the coreference array $ A = \{m_1, 0, m_1, m_2,0, m_2, m_3,0, m_3, m_1...\}$, where $m_n$ denotes the nth mention cluster, and 0 denotes no mentions, the enriched attention for additive attention fusing is formulated as:
\begin{equation}
\begin{aligned}
& Attention(Q, K, V) = Softmax(\frac{QK^T}{\sqrt{d_k }} + M_A)V, \\
& head_i = Attention(QW_i^Q, KW_i^K, VW_i^V) ,
\end{aligned}
\end{equation}
where $M_A$ is a coreference matrix constructed from the coreference array $A$ with the element value in the matrix calculated by adding (for additive model) or multiplying (for multiplication model) the coreference hyper-parameter $coref_{weight}$ with the original attention weight if the element belongs to the coreference array, $Q, K, V$ are the query, key and value respectively, $d_k$ is the dimension of the keys, and $W_i$ is trainable parameter. 
For dot product (multiplication) fusing, it is formulated as:
\begin{equation}
\begin{aligned}
& Attention(Q, K, V) =  Softmax(\frac{QK^T}{\sqrt{d_k}} \odot M_A)V, \\
& head_i = Attention(QW_i^Q, KW_i^K, VW_i^V) ,
\end{aligned}
\end{equation}
where we calculate the dot product of  $\frac{QK^T}{\sqrt{d_k}} $ and a coreference matrix $M_A$ constructed from the coreference array $A$.


\subsection{Integration}
A machine reading comprehension task expects the model to output the start and end positions of the answer. For the RCGN method, we fuse the hidden state of nodes $v_i$ in the last layer of RCGN and the embeddings from the pre-trained language model with a fully-connected (FC) layer ,  and then calculate the start/end positions of the answer. 
\begin{equation}
\begin{aligned}
& E = FC(E_{prLM}||E_{gnn}), \\
& P_s = argmax(softmax(W_sS)),
\end{aligned}
\end{equation}
where $E_{prLM}$ denotes the embeddings from the pre-trained language model, $E_{gnn}$ denotes the embeddings from the graph encoder,  $P_s$ denotes the predicted start positions, $W_s$ denotes the weight matrix and $S$ denotes the text feature.

For the two methods that add one additional encoder layer for additive or multiplication attention enrichment, we directly used the output of that encoder layer for the follow-up processing.

Following the practice of CorefRoBERTa \cite{ye-etal-2020-coreferential} in handling multiple answers for the same question, we use the cross entropy to calculate the losses for each answer if the question has multiple answers:
\begin{equation}
\begin{aligned}
& E_n = FC(E_{prLM}, n), \\
& L_s = \sum_i^nH(p_si,q_si), \\
& L_e = \sum_i^nH(p_ei,q_ei), \\
& L_{total} = avg(L_s + L_e + L(E_n, n)),
\end{aligned}
\end{equation}
where $n$ denotes the answer count as a hyper parameter for handling multiple answers, $E_n$ denotes the results after the linear transformation of the embeddings for the answer count and then we obtains the predicted start positions and end positions from that embeddings, $L({E_n}, n)$ denotes the cross-entropy loss between the transformed embeddings and the answer count,  $L_s$ denotes the total loss of the start positions, $L_e$ denotes the total loss of the end positions and $L_{total}$ denotes the combined total loss.

\section{Experiments}
\subsection{Model Settings}
We developed three models based on the sequence-to-sequence Transformer architecture. The pre-trained RoBERTa-large was used as the base model and then we used the following three methods to fine-tune it: 1) $\rm Coref_{GNN} $: feeding the coreference information into a graph neural network and then fuse the representations; 2)  $\rm Coref_{AddAtt}$: adding the coreference weights with the self-attention weights; 3) $\rm Coref_{MultiAtt}$: calculating the dot product of the coreference weights with the self-attention weights. We used the results from CorefRoBERTa \cite{ye-etal-2020-coreferential} as our baselines.

\subsection{Setup}
Our coreference resolution was implemented in spaCy \cite{spacy2} and NeuralCoref. NeuralCoref is an extension for spaCy that is trained on the OntoNotes 5.0 dataset based on the training process proposed by Clark and Manning \cite{clark-manning-2016-deep}, which identifies the coreference clusters in the text as mentions. In particular, spaCy 2.1.0 and NeuralCoref 4.0 are used, because the latest spaCy version 3.0+ has compatibility issues with NeuralCoref and extra efforts are required to solve the issues. 

The neural network implementation was implemented in PyTorch \cite{NEURIPS2019_bdbca288} and Hugging Face Transformers \cite{wolf-etal-2020-transformers}. We used the embeddings of the pre-trained language model $\rm RoBERTa_{LARGE}$, with the relational graph convolutional network implemented in Deep Graph Library (DGL) \cite{wang2020deep}. We used Adam \cite{kingma2017adam} as our optimizer, and the learning-rate was \{1e-5, 2e-5, 3e-5\}. We trained each model for \{4, 6\} epochs and selected the best checkpoints on the development dataset with Exact match and F1 scores. All experiments were run on two NVIDIA TITAN RTX GPUs, each with 24GB memory.
\subsection{Tasks and Datasets}
Our evaluation was performed on the QUOREF dataset \cite{Dasigi2019Quoref}. The dataset contains a train set with 3,771 paragraphs and 19,399 questions, a validation set with 454 paragraphs and 2,418 questions, and a test set with 477 paragraphs and 2,537 questions.
\subsection{Results}
We quantitatively evaluated the three methods and reported the standard metrics: exact match score (EM) and word-level F1-score (F1) \cite{rajpurkar-etal-2016-squad}. 
\begin{table}
\small\setlength{\tabcolsep}{5pt}
{
\begin{tabular}{l c c c c }
\toprule
\multicolumn{1}{c}{ \textbf{Model}}             & \multicolumn{2}{c}{ \textbf{Dev}} & \multicolumn{2}{c}{ \textbf{Test}} \\

                  & \textbf{EM}         & \textbf{F1}         & \textbf{EM}          & \textbf{F1}         \\
\midrule
 $\rm QANet^*$             & 34.41      & 38.26      & 34.17       & 38.90      \\
 QANet + $\rm BERT_{BASE}^*$ & 43.09      & 47.38      & 42.41       & 47.20 \\
 $\rm BERT_{BASE}^+$    & 61.29   & 67.25    & 61.37   & 68.56 \\
 $\rm CorefBERT_{BASE}^+$  & 66.87  & 72.27   & 66.22 & 72.96 \\
 \midrule
 $\rm BERT_{LARGE}^+$    & 67.91   & 73.82    & 67.24   & 74.00 \\
 $\rm CorefBERT_{LARGE}^+$    & 70.89   & 76.56    & 70.67   & 76.89 \\
 \midrule
 $\rm RoBERTa_{LARGE}^+$    & 74.15   & 81.05    & 75.56   & 82.11 \\
 $\rm CorefRoBERTa_{LARGE}^+$    & 74.94   & 81.71    & 75.80   & 82.81 \\
 \midrule
 $\rm Coref_{GNN}$  & 79.23   & 85.89    & 78.60   & 85.15 \\
 $\rm Coref_{AddAtt}$  & \textbf{80.02}   & \textbf{86.13}    & \textbf{79.11}   & \textbf{85.86} \\
 $\rm Coref_{MultiAtt}$  & 79.85   & 86.02    & 78.52   & 85.27 \\
 \bottomrule
\end{tabular}}
\caption{Exact Match and F1 scores of baselines and our proposed models. Results with *, + are from \citet{Dasigi2019Quoref} and \citet{ye-etal-2020-coreferential} respectively.}\label{tab1}
\end{table}

As shown in Table \ref{tab1}, compared with the baseline model CorefRoBERTa, the performance of our models improves significantly. In particular, $\rm Coref_{AddAtt}$ performs best with 5.08\%, 4.42\% improvements over the baseline model in Exact Match and F1 score respectively on the QUOREF dev set, and 3.05\% (F1) and 3.31\% (Exact Match) improvements on the QUOREF test set. $\rm Coref_{GNN}$ and $\rm Coref_{MultiAtt}$ also outperform the baseline model by 2.34\% (F1) and 2.80\% (Exact Match), and 2.46\% (F1) and 2.72\% (Exact Match) respectively on the test set. Compared with the $\rm RoBERTa_{LARGE}$ that does not use any explicit coreference information in the training or the $\rm CorefRoBERTa_{LARGE}$  that uses the coreference information in the training, the improvements of our model are higher, which proves the effectiveness of the explicit coreference instructions in our strategies.

\newcommand{\tabincell}[2]{
\begin{tabular}{@{}#1@{}}#2\end{tabular}
}

\begin{table*}[htb]
\begin{small}
\centering
\begin{tabular}{p{0.63\linewidth}p{0.13\linewidth}p{0.17\linewidth}  }
\toprule
 \textbf{Coref-resolved Context (Abbreviated)}   & \textbf{Question}  & \textbf{Answers}\\
\midrule
\textit {\textcolor{blue}{\textbf{Henrietta}} take an immediate liking to her, and she asks if Luce can sit by  {\textcolor{blue}{\textbf{her}}} during the wedding. Rachel arrives with her father and the ceremony begins. As Rachel is walking down the aisle, her eyes wander and she makes eye contact with Luce.}   &  \textit {Rachel makes eye contact with a woman sitting next to whom?} &   \textit{Henrietta}  (Golden)\break \textcolor{red}{Rachel} \hspace{0.6cm}(CorefR)\break   Henrietta \hspace{0.1cm}($\rm C_{AddAtt}$)  \\
\midrule
\textit {After the song was completed, they wanted to play it to \textcolor{blue}{\textbf{Rihanna}}, but Blanco was skeptical about the reaction towards the song because of its slow sound. After StarGate played it to her, they called Blanco from London and told him that \textcolor{blue}{\textbf{she}} liked the song: ``She's flippin' out.}  & \textit{Who liked a song?} & \textit{Rihanna} (Golden) \textcolor{red}{Blanco} (CorefR)    Rihanna \hspace{0.21cm}  ($\rm C_{AddAtt}$) \\
\bottomrule
\end{tabular}
\caption{Comparison of the predictions for two questions in QUOREF dev set. The blue and bold words indicate the mentions in the same coreference cluster obtained from coreference resolution. In the Answers column, Golden indicates the golden answer; CorefR indicates the prediction made by $ \rm CorefRoBERTa_{LARGE}$ model; $\rm C_{AddAtt}$ indicates the prediction made by $\rm Coref_{AddAtt}$ model.} \label{tab3}
\end{small}
\end{table*}



\begin{figure*}[!htbp]
  \centering
  \includegraphics[width=\textwidth]{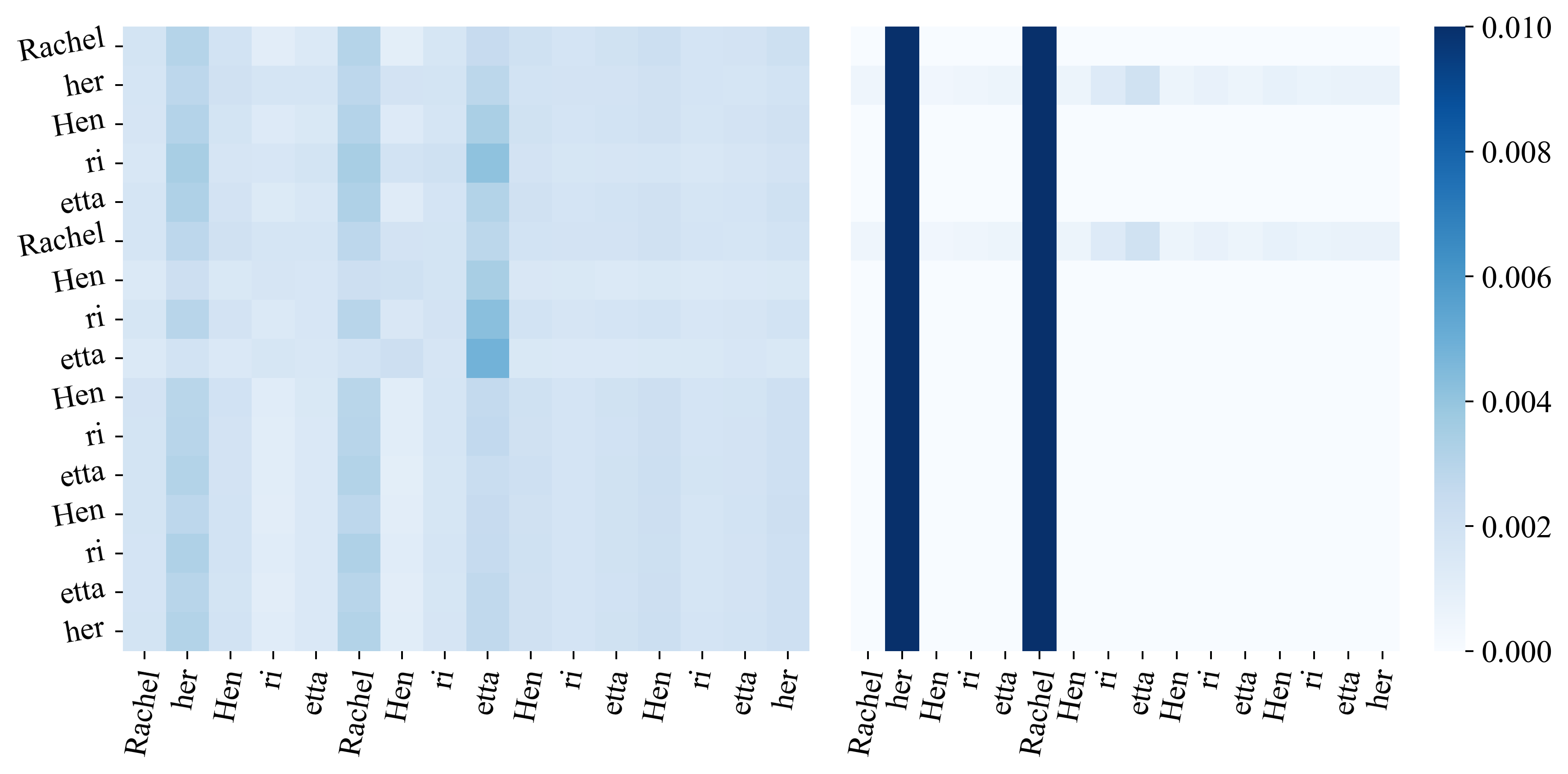}
  \caption{Sample average cross attentions for all heads from $\rm Coref_{AddAtt}$ model (left) and $\rm CorefRoBERTa_{LARGE}$ model (right).  The cross attentions among the anaphoric expressions and the entities of our model ($\rm Coref_{AddAtt}$ ) are visibly much more distinctive than those of the baseline model ($\rm CorefRoBERTa_{LARGE}$ ). }
  \label{fig:corefAdditive}
\end{figure*}
\section{Analysis}
\subsection{Model Efficiency}
As shown in Table 2, compared with $\rm RoBERTa_{LARGE}$, our methods added only one component that explicitly incorporates the coreference information, and the three methods we used all exhibit considerable improvements over the baselines. Compared with $\rm RoBERTa_{LARGE}$  which has 354M parameters,  $\rm Coref_{AddAtt}$ and the  $\rm Coref_{MultiAtt}$ add an encoder layer, which adds over 12M parameters. For the $\rm Coref_{GNN}$ method, we added one hidden layer in GNN and two linear layers to transform the feature dimensions, with around 68.7K parameters in total.
Our predictions are that intuitively with more focuses on the coreference clues, the models perform better on the task that requires intensive coreference resolution, as we have explicitly increased the attention weights to connect the words in the same coreference mention clusters. However, the overall performance of the models is also limited by the performance of the coreference component we use, namely, NeuralCoref.
\begin{table*}[htb]
\centering
\begin{small}
\begin{tabular}{p{0.52\linewidth}p{0.18\linewidth}p{0.20\linewidth}}
\toprule
\textbf{Coref-resolved Context (Abbreviated)}   & \textbf{Question}  & \textbf{Answers}\\
\midrule
\textit{West Point cadet \textcolor{blue}{\textbf{Rockwell "Rocky" Gilman}} is called before a hearing brought after an influential cadet, Raymond Denmore, Jr., is forced to leave the academy...Denmore's attorney, Lew Proctor, attacking \textcolor{red}{\textbf{the academy}} and \textcolor{red}{\textbf{its}} Honor Code system, declares that \textcolor{blue}{\textbf{Gilman}} is unfit and possibly criminally liable.}  &  \textit{Who's honor code system does Proctor attack?} & \textit{the academy} (Golden)  \textcolor{red}{West Point} \hspace{0.35cm} ($\rm C_{AddAtt}$)   \\
\midrule
\textit{Following a career hiatus that reignited \textcolor{magenta}{\textbf{her}} creativity, \textcolor{blue}{\textbf{Beyoncé}} was inspired to create a record with a basis in traditional rhythm and blues that stood apart from contemporary popular music...Severing professional ties with \textcolor{red}{\textbf{father}} and manager \textcolor{red}{\textbf{Mathew Knowles}}, \textcolor{blue}{\textbf{Beyoncé}} eschewed the music of \textcolor{magenta}{\textbf{her}} previous releases}  & \textit{What is the last name of the person who went on a career hiatus?} &  \textit{Knowles} (Golden) \hspace{6.4cm} \textcolor{red}{Beyoncé} \hspace{0.7cm}($\rm C_{AddAtt}$) \\
\midrule
\textit{When the prosecutor suggests that the crime would have still happened if the owner were a woman, \textcolor{blue}{\textbf{Christine, Andrea}}, \textcolor{red}{\textbf{Annie}}, Janine and the other women who witnessed the crime all laugh and exit the courtroom.}  & \textit{What are the names of the women Janine has to determine are sane or crazy?} &  \textit{Christine, Andrea, Annie} \break (Golden) \break \textcolor{red}{Christine, Andrea}\hspace{0.5cm} \break ($\rm C_{AddAtt}$) \\
\bottomrule
\end{tabular}
\caption{Errors in predictions for three questions in QUOREF dev set. The blue and bold words indicate the mentions in the same coreference cluster. The bold words in red or magenta indicate the failure of our model in making necessary reasoning.  In the Answers column, Golden indicates the golden answer; $\rm C_{AddAtt}$ indicates the prediction made by $\rm Coref_{AddAtt}$ model.} \label{tab4}
\end{small}
\end{table*}

\subsection{Case Studies}
To understand the model's performance beyond the automated metrics, we analyze our predicted answers qualitatively.
Table \ref{tab3} compares the representative answers predicted by our models and $\rm CorefRoBERTa_{LARGE}$. These examples require that the models should precisely locate the entity from several distracting entities for the anaphoric expression that directly answers the questions. Our model demonstrates that, after resolving the anaphoric expression with the antecedents in the context and enhancing with the coreference information by connecting the anaphoric expression with its antecedents, such as the connection from \textit{\textbf{her}} to \textit{\textbf{Henrietta}} in the first example and the connection from \textit{\textbf{she}} to \textit{\textbf{Rihanna}} in the second example, our model accurately locates the entity name among several names in the context, which the $\rm CorefRoBERTa_{LARGE}$ fails to uncover.

We further explored the effects of the anaphoric connections on the attention weights by comparing the attention weights of the sample in the first row in Table \ref{tab3} between our $\rm Coref_{AddAtt}$ and $\rm CorefRoBERTa_{LARGE}$ model, as shown in Figure \ref{fig:corefAdditive}. It is clear that the anaphoric expressions are not connected in the $\rm CorefRoBERTa_{LARGE}$ model, as indicated by the obtrusive attentions on \textbf{Rachel} and \textbf{Her} in the heatmap on the right of the figure. For the  $\rm Coref_{AddAtt}$, the varying colors on the left heat-map indicate the connection strength among the anaphoric expressions and evidence the effects of explicit coreference addition that smooth and strength the attentions for anaphoric expressions, which contributes to the higher performance of our models.

\subsection{Error Analysis}
Despite the improvements made by our model, it still fails to predict the correct answers for some questions. We analyzed and summarized several error cases as follows.

Table \ref{tab4} shows three representative types of errors. The first type of errors is caused by the limitations of the coreference resolution component, NeuralCoref, as its performance had not reached 80\% in F1 for MUC, $\rm B^3$ or $\rm CEAF_{\phi4}$ \cite{clark-manning-2016-deep}, which is evidenced by the failure in resolving the antecedent of the anaphoric expression \textit{\textbf{its}} as \textit{\textbf{the academy}} in the first sample, and the failure in clustering the anaphoric expressions \textit{\textbf{her}} with the entity \textit{\textbf{Beyoncé}} in the second sample,  despite the success in resolving the second \textit{\textbf{Gilman}} to its antecedent \textit{\textbf{Rockwell "Rocky" Gilman}}. The second type of errors is more complicated, which involves multi-step reasoning that cannot be handled by simply adding the coreference information. To correctly answer the second question, the model should perform two successive tasks successfully: 1) it should understand that \textit{\textbf{Mathew Knowles}} is the father of \textit{\textbf{Beyoncé}}; 2) it should understand the world knowledge that the last name of \textit{\textbf{Beyoncé}} is the same as her father's, which should be \textit{\textbf{Knowles}}. This type of errors shows that our model performs poorly on the questions that require multi-step reasoning. The third type of errors is caused by the questions that have multiple items in an answer. A hyperparameter that limits the total number of items in an answer is used in our models and this parameter is set to 2 in the training, thus when the number of total items in the answer exceeds 2, our models fail to predict the exact items, and the third item \textit{\textbf{Annie}} is ignored.

\section{Conclusion}
In this paper, we present intuitive methods to solve coreference-intensive machine reading comprehension tasks by following the reading process of human in which people connect the anaphoric expressions with explicit instructions. We demonstrate that all our three fine-tuning methods, including $\rm Coref_{GNN}$,  $\rm Coref_{AddAtt}$ and  $\rm Coref_{MultiAtt}$,   are superior to the pre-trained language models that incorporate the coreference information in the pre-training stage, such as $ \rm CorefRoBERTa_{LARGE}$. As the fine-tuning methods rely on the coreference resolution models supplied by other researchers, their performance is also constrained by the accuracy of those coreference resolution models. In addition, the questions that require multi-step reasoning, span multiple entities or contain multiple answer items also pose the challenges to our models. In the future, with more in-depth study on human reasoning in reading comprehension and more progress in graph neural networks, the GNN-based coreference graph can be enriched with more edge types and diverse structures to leverage more linguistic knowledge and gain better performance.

\section*{Acknowledgement}
We would like to thank Yuchen He for the help of this work.  We also appreciate the valuable feedback from the anonymous reviewers.

\bibliography{acl}
\bibliographystyle{acl_natbib}

\end{document}